\documentclass[letterpaper, 10 pt, conference]{ieeeconf} 
\IEEEoverridecommandlockouts  
\usepackage[T1]{fontenc}
\usepackage{amsmath,amsfonts}
\usepackage{algorithmic}
\usepackage{textcomp}
\usepackage{stfloats}
\usepackage{url}
\usepackage{verbatim}
\usepackage{graphicx}
\usepackage{cite}
\usepackage{bm}
\usepackage[colorlinks, linkcolor=black, anchorcolor=black, citecolor=black]{hyperref}
\usepackage{flushend}
\usepackage{amsmath}
\usepackage{subcaption}
\usepackage{booktabs}
\usepackage{float}
\usepackage{afterpage}
\begin{document}
\title{MASQ: Multi-Agent Reinforcement Learning for \\
Single Quadruped Robot Locomotion}

\author{
    Qi Liu$^\dagger$, Jingxiang Guo$^\dagger$, Sixu Lin, Shuaikang Ma, Jinxuan Zhu, Yanjie Li\textsuperscript{*}
    \thanks{This work was supported by the National Natural Science Foundation of China [61977019, U1813206] and Shenzhen Fundamental Research [JCYJ20220818102415033, JSGG20201103093802006, KJZD20230923114222045]. \textit{(Corresponding author: Yanjie Li, autolyj@hit.edu.cn)}}
    \thanks{The authors are with the Guangdong Key Laboratory of Intelligent Morphing Mechanisms and Adaptive Robotics and the School of Mechanical Engineering and Automation, the Harbin Institute of Technology Shenzhen, 518055, China.}
    \thanks{\textsuperscript{*} Corresponding author.}
    \thanks{$^\dagger$ Equal contribution.}
}


\maketitle
\begin{abstract}
This paper proposes a novel method to improve locomotion for a single quadruped robot using multi-agent deep reinforcement learning (MARL). Many existing methods use single-agent reinforcement learning for an individual robot or MARL for the cooperative task in multi-robot systems. Unlike existing methods, this paper proposes using MARL for the locomotion of a single quadruped robot. We propose a learning structure called Multi-Agent Reinforcement Learning for Single Quadruped Robot Locomotion (MASQ),  considering each leg as an agent to explore the action space of the quadruped robot, sharing a global critic, and learning cooperatively. Experimental results show that MASQ not only speeds up learning convergence but also enhances robustness in real-world settings, suggesting that applying MASQ to single robots such as quadrupeds could surpass traditional single-robot reinforcement learning approaches. Our study provides insightful guidance on integrating MARL with single-robot locomotion.
\end{abstract}


\section{Introduction}
\label{Section: Introduction}

Reinforcement learning (RL) has made remarkable progress in various robot control learning \cite{10610833}, such as quadruped robot \cite{margolis2022walk,10161144}, biped robot \cite{benbrahim1997biped,10611621}, and unmanned aerial vehicle \cite{8004441}. This paper focuses on the quadruped robot control learning. 

In the process of applying deep RL to a single robot, it is prevalent to use single-agent algorithms\cite{10610286,lobbezoo2021reinforcement,10011921,10341908}. However, single-agent algorithms may have limitations in managing coordination in specific problems. Many existing methods use single-agent RL algorithms for individual robot learning or multi-agent deep reinforcement learning (MARL) for multi-robot systems in cooperative tasks \cite{canese2021multi,orr2023multi,chen2022towards}. Cooperative MARL algorithms have been widely demonstrated in game-AI \cite{rashid2020monotonic,9248616} to have advantages in multi-agent cooperation. 

\begin{figure}[htbp]
    \centering
    \subfloat[Dog on grass]{  
    \begin{minipage}[b]{0.24\textwidth} 
        \centering
        \includegraphics[width=4cm,height=3cm]{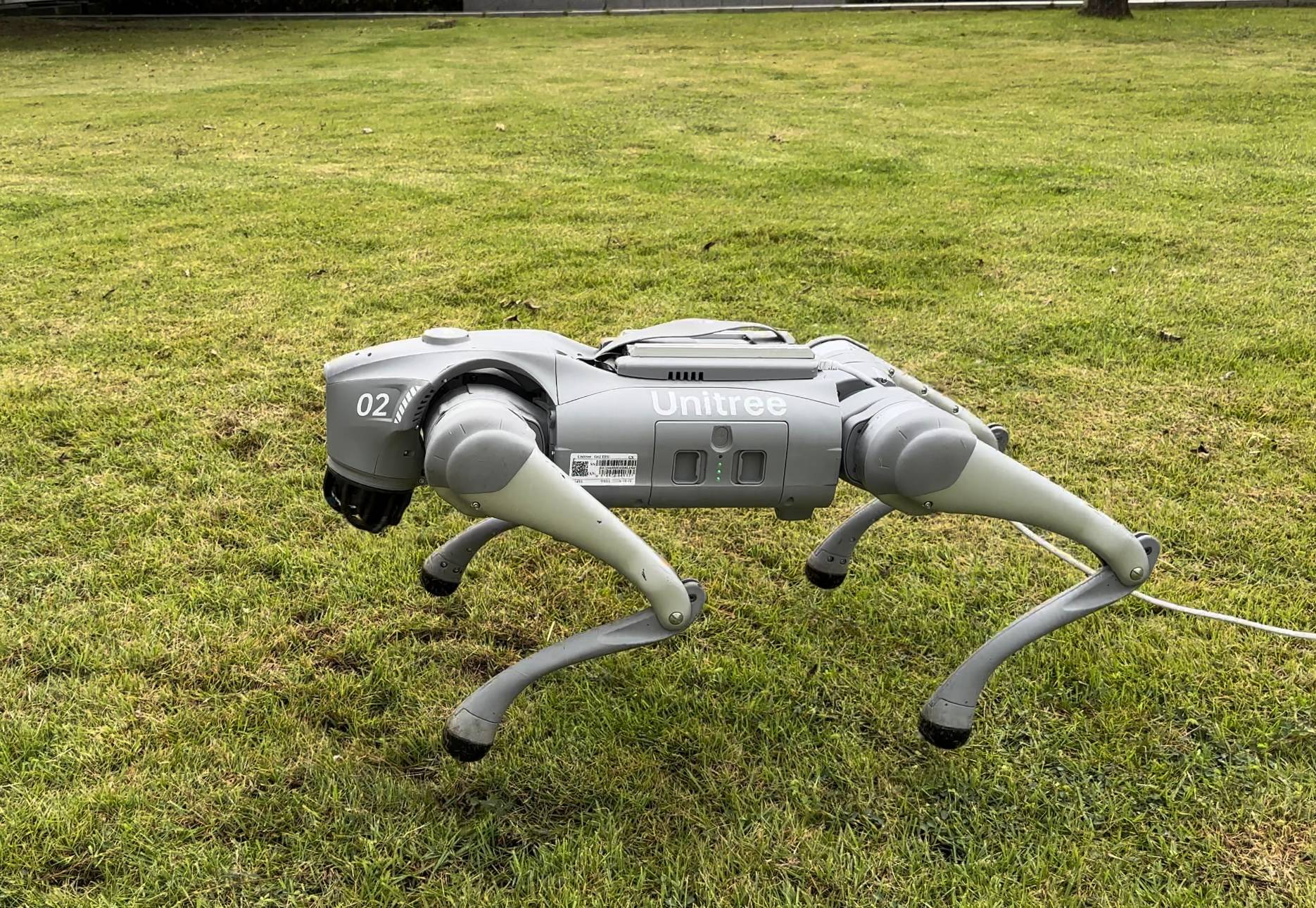} 
    \end{minipage}}
    \subfloat[Dog on rock]{ 
    \begin{minipage}[b]{0.24\textwidth} 
        \centering
        \includegraphics[width=4cm,height=3cm]{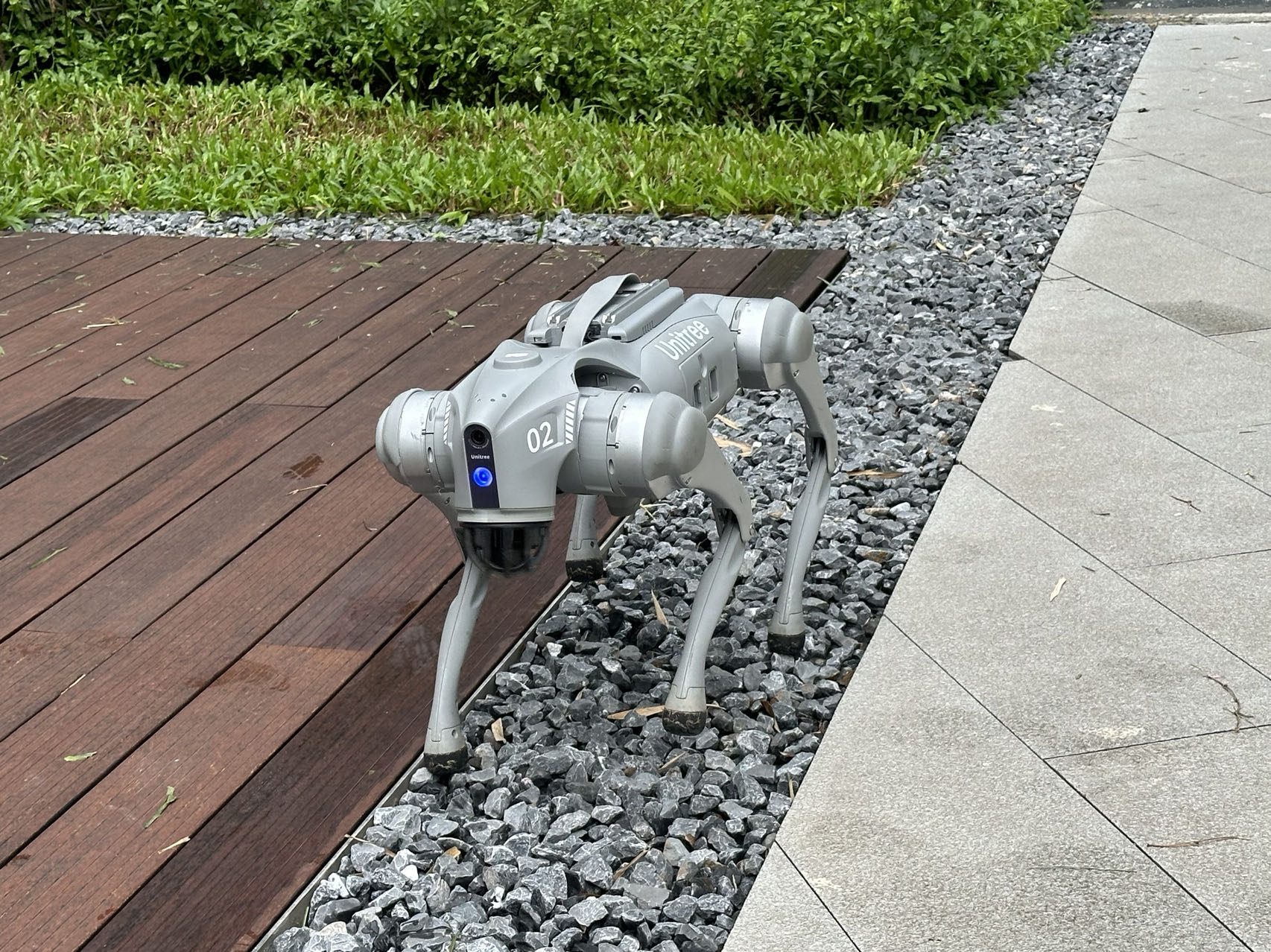} 
    \end{minipage}}

    \subfloat[Dog on flat]{
    \begin{minipage}[b]{0.24\textwidth} 
        \centering
        \includegraphics[width=4cm,height=3cm]{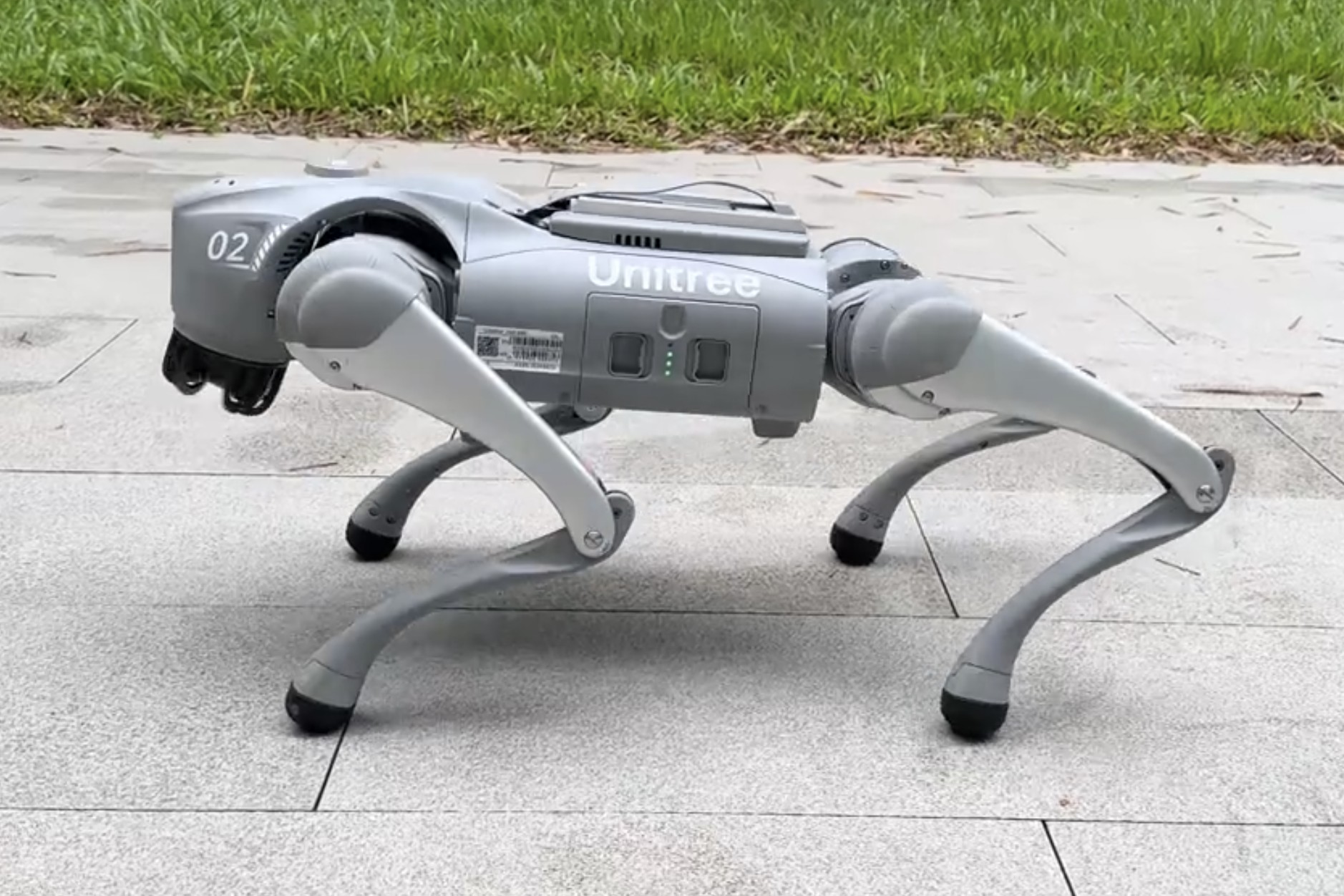} 
    \end{minipage}}
    \subfloat[Dog on rubber track]{
    \begin{minipage}[b]{0.24\textwidth} 
        \centering
        \includegraphics[width=4cm,height=3cm]{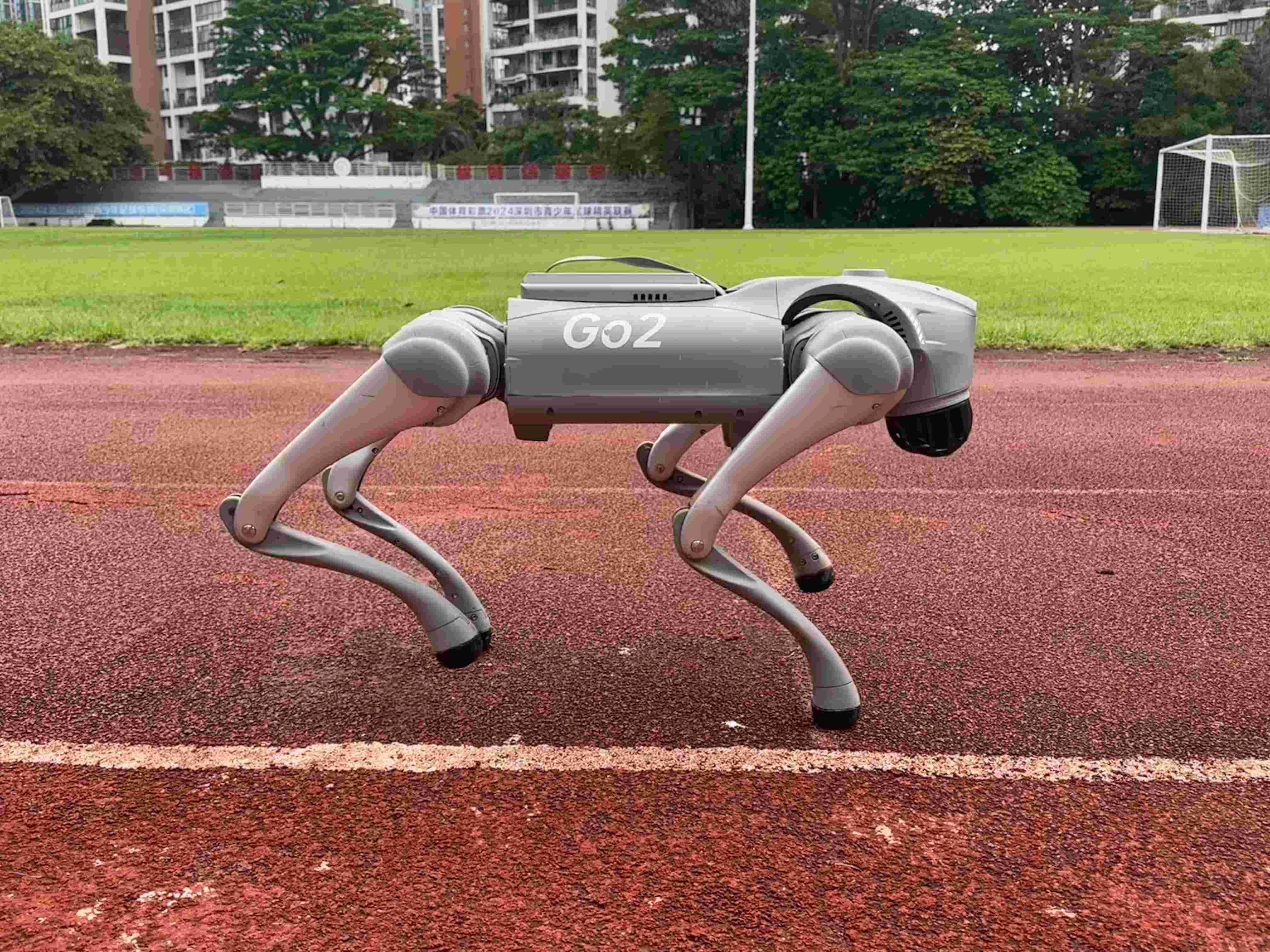} 
    \end{minipage}}
    \caption{Quadruped robot on various terrains}
    \label{Figure: dog_pictures} 
\end{figure}

\begin{figure*}[!htbp]
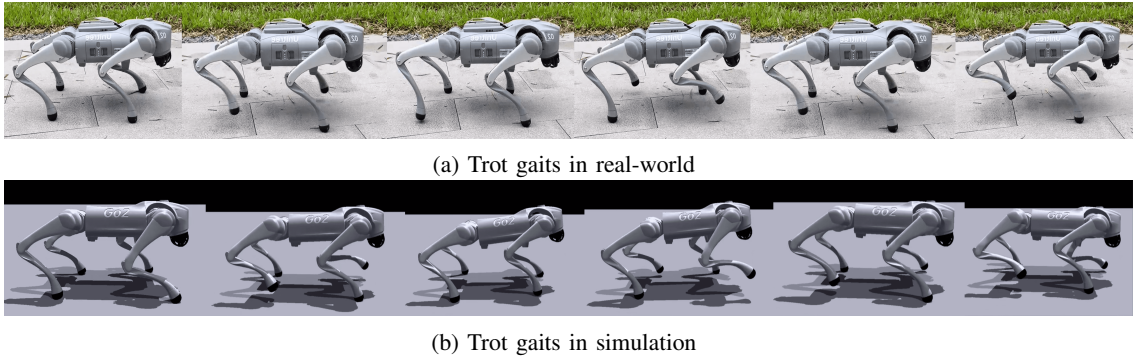
 
    \centering
    \subfloat[Trot gaits in real-world]{  
    \begin{minipage}[b]{0.9\textwidth} 
        \centering
        \includegraphics[width=15cm,height=1.8cm]{pace_list_compressed.png} 
    \end{minipage}}

    \subfloat[Trot gaits in simulation]{ 
    \begin{minipage}[b]{0.9\textwidth} 
        \centering
        \includegraphics[width=15cm,height=1.8cm]{pacesim_list_compressed.png} 
    \end{minipage}}
    \caption{Sim-to-Real comparison of trot gaits}
    \label{Figure: Sim-to-Real Comparison of Trot Gaits} 
\end{figure*}

Unlike existing approaches, this paper proposes using MARL for the locomotion of a single quadruped robot to enhance cooperation between its four legs, thereby enabling it to navigate complex terrains and perform intricate tasks. By proposing cooperative MARL, where each leg acts as an agent, the quadruped robot can better coordinate its movements. This collaborative learning structure, termed Multi-Agent Reinforcement Learning for Single Quadruped Robot Locomotion (MASQ), allows the robot to share experiences. Fig. \ref{Figure: dog_pictures} presents the deployment of the MASQ algorithm on a quadruped robot tested across different terrains, including grass, rocks, flat surfaces, and rubber tracks, demonstrating its performance in diverse environments. Fig. \ref{Figure: Sim-to-Real Comparison of Trot Gaits} 
shows a sim-to-real comparison of trot gaits in a quadruped robot, highlighting the consistency between simulated and real-world gait patterns.

The main contributions of this paper can be summarized as follows:
\begin{itemize}
    \item This paper proposes MASQ, a method that treats each leg of a quadruped robot as an individual agent. The locomotion learning is modeled as a cooperative multi-agent reinforcement learning (MARL) problem and solved using a MARL algorithm.
    \item Experimental results show that the proposed method enhances performance in executing gaits, improves training efficiency and robustness, and achieves better final performance, demonstrating the value and impact of the approach.
\end{itemize}

\section{Related Work}
\label{Section: Related work}

\subsection{Deep RL for Single Robot Control}
\label{Subsection: deep RL for quadruped robot control}

Recent advances in deep RL for quadruped robots are driven by simulation technologies such as Isaac Gym \cite{makoviychuk2021,rudin2022learning}. Hardware advancements have shown robust performance on various tasks through Sim2Real transfer with zero shot \cite{Hwangbo_2019,doi:10.1126/scirobotics.abc5986,doi:10.1126/scirobotics.abk2822,gangapurwala2021real}. Current research focuses on task-specific reward composition and training paradigms to bridge the Sim2Real gap \cite{kumar2021rma,9981091,choi2023learning}, often using Proximal Policy Optimization (PPO) \cite{schulman2017proximal} with an emphasis on task and reward design rather than novel RL algorithms. Although some studies have explored learning algorithms to improve efficiency \cite{kim2024not,ye2023knowing,semage2023,gu2024, Peng_2018,9205217}, most efforts remain centered on efficiency rather than modifying algorithms to control the characteristics of the object. Additionally, research has delved into task-specific strategies like rapid motor adaptation (RMA) \cite{kumar2021rma}, hierarchical RL for multi-skill tasks \cite{Zhou2022}, and symmetry-based data enhancement \cite{zhang2023,mittal2024symmetry}, yet challenges remain in using robot symmetry through single-agent methods, indicating that algorithmic research on this aspect is still underdeveloped.

\subsection{MARL for Multi-robot Control}
\label{Subsection: MARL for multi-robot control}

In multi-agent settings, algorithms like Multi-Agent Proximal Policy Optimization (MAPPO) \cite{yu2022surprising}, Temporally Extended Multi-Agent Reinforcement Learning (TEMP) \cite{machado2023temporal} have demonstrated strong capabilities in addressing multi-agent robotic challenges, such as drone fleet control \cite{alon2020multiagent,park2022coordinated} and autonomous vehicle fleets \cite{sainzpalacios2022deep,schmidt2023learning}. Methods like Reinforced Inter-Agent Learning (RIAL) and Differentiable Inter-Agent Learning (DIAL) \cite{foerster2016learning} further enhance collaborative performance by developing communication protocols among agents. This paper proposes modeling single-quadruped robot locomotion as a cooperative MARL problem, where each leg is treated as an independent agent, contrasting with previous approaches that treat the robot as a monolithic entity \cite{lifelike_agility,wang2024learning,han2024learning,review_quadrupedal_rl} or cooperative groups of multi-robots \cite{orr2023multi,lan2021towards,10035491}.

\section{Preliminaries}
\label{Section: Preliminary}

This paper considers a finite-horizon Markov decision process (MDP) \cite{sutton2018reinforcement}, defined by a tuple $(\mathcal{S}, \mathcal{A}, \mathcal{P}, r, \gamma, T)$. $\mathcal{S}$ denotes the state space, $\mathcal{A} := \{a_0, a_1, ..., a_{|\mathcal{A}|-1}\}$ represents a finite action space, $\mathcal{P}: \mathcal{S} \times \mathcal{A} \times \mathcal{S} \rightarrow [0,1]$ represents the staåte transition distribution, $r: \mathcal{S} \times \mathcal{A} \rightarrow \mathbb{R}$ denotes a reward function, $\gamma \in [0,1)$ denotes a discount factor, and $T$ is a time horizon. At each time step $t$, the policy $\pi$ selects an action $a_{t} \in \mathcal{A}$. After entering the next state by sampling from $\mathcal{P}\left(s_{t+1} \mid s_{t}, a_{t}\right)$, the agent receives an immediate reward $r\left(s_{t}, a_{t}\right)$. The agent continues to perform actions until it enters a terminal state or $t$ reaches the time horizon $T$. RL aims to learn a policy $\pi: \mathcal{S} \times \mathcal{A} \rightarrow [0, 1]$ for decision-making problems by maximizing discounted rewards. For any policy $\pi$, the state-action value function ($Q$ function) is defined as
\begin{equation}
    Q^{\pi}(s, a)={\mathbb{E}^{\pi}}\left[\sum_{t=0}^{T} \gamma^{t} r\left(s_{t}, a_{t}\right) \mid s_{0}=s, a_{0}=a\right]
    \label{Eq: Q function}
\end{equation}

Proximal Policy Optimization (PPO) \cite{schulman2017proximal} enhances the stability and performance of policy gradient methods by limiting policy updates to prevent destabilizing deviations. The core PPO update rule optimizes a clipped surrogate function:
\begin{equation}
    L^{\textit{CLIP}}(\theta) = \mathbb{E}_t \left[ \min \left( r_t(\theta) \hat{A}_t, \text{clip}(r_t(\theta), 1 - \epsilon, 1 + \epsilon) \hat{A}_t \right) \right]
\end{equation}
where
\begin{equation}
    r_t(\theta) = \frac{\pi_{\theta}(a_t | s_t)}{\pi_{\theta_{\text{old}}}(a_t | s_t)}
\end{equation}
and
\begin{equation}
    \hat{A}_t = Q^{\pi}(s_t, a_t) - V_{\psi}(s_t)
\end{equation}
where $\psi$ denotes the parameters of value function ($V_{\psi}$) network, $\epsilon$ denotes a coefficient. Policy parameters $\theta$ are updated as:
\begin{equation}
    \theta \leftarrow \theta + \alpha \nabla_{\theta} L^{\textit{CLIP}}(\theta)
\end{equation}
PPO's constrained updates stabilize training and improve performance, making it practical for complex single-agent RL tasks.

\section{Multi-Agent Reinforcement Learning for Single Quadruped Robot Locomotion}
\label{Section: Multi-Agent Reinforcement Learning for Single Quadruped Robot Locomotion}

\begin{figure*}[htbp]
    \centerline{\includegraphics[width=15cm, height=14cm]{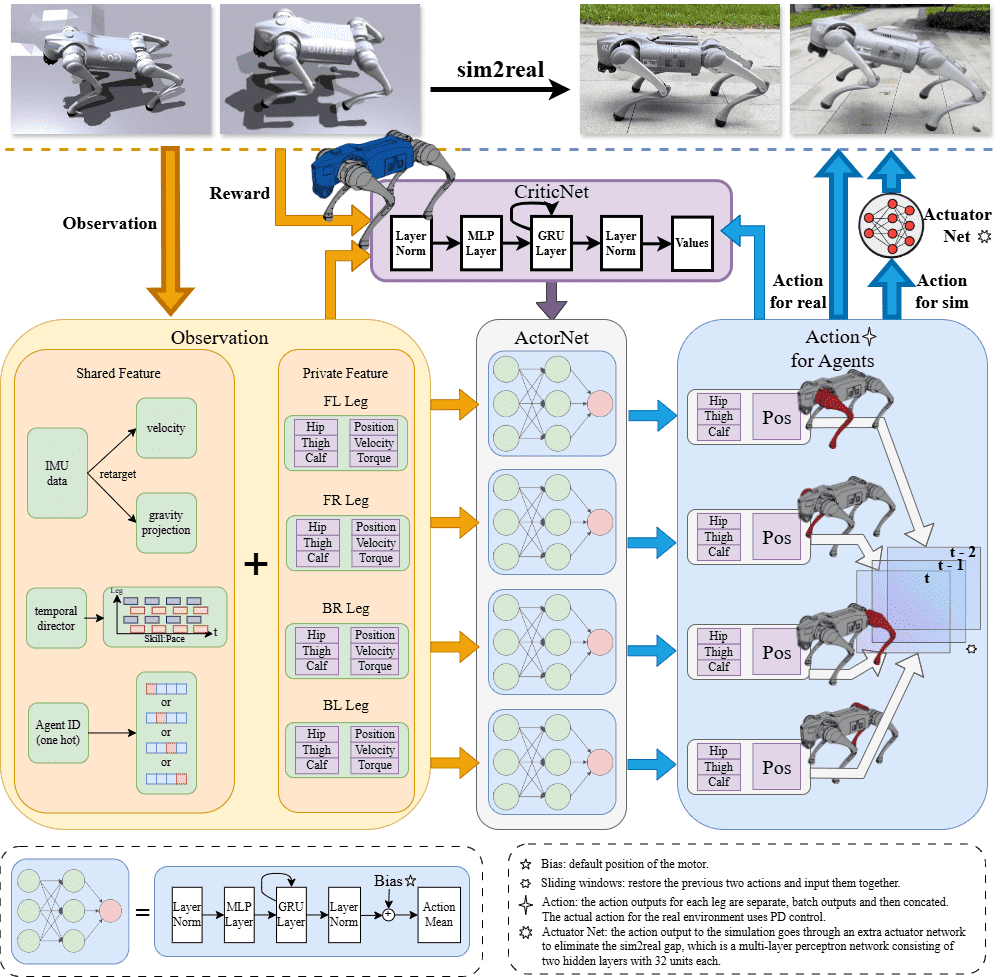}}
    \caption{The framework of MASQ}
    \label{Figure: learning process}
\end{figure*}

In this paper, we use the collaborative potential of multiple agents to improve the learning process for a single robot's locomotion, resulting in faster training convergence, robustness and better final performance in real-world environments. Specifically, each leg of the quadruped robot is treated as a separate agent within the multi-agent structure, with individual observations and a shared global critic, significantly improving the cooperation among the robot's limbs for more effective locomotion.

\subsection{MASQ Modeling}
\label{Subsection: MASQ Modeling}

This paper models a single quadruped robot locomotion as a cooperative multi-agent problem, which is described as a partially observable decentralized Markov decision process (decPOMDP) \cite{ong2009pomdps}. The decPOMDP is defined by the tuple $G=(\mathcal{S}, \mathcal{A}, \mathcal{P}, r, \mathcal{Z}, \mathcal{O}, N, \gamma, T)$. $\mathcal{S}$ is the state space, $\mathcal{A}$ is the action space, $\mathcal{P}$ is the state transition distribution, $r$ is the reward function, $\mathcal{Z}$ is the observation space, $\mathcal{O}$ is the observation function, $N$ is the number of agents, $\gamma$ is the discount factor, and $T$ is the time horizon. At each time step $t$, each agent $n \in \{1, \ldots, N\}$ selects an action $a^n_t \in \mathcal{A}$, resulting in a joint action $\bm{a_t} = \{a^1_t, a^2_t, \ldots, a^N_t\}$. The environment transitions to a new state $s_{t+1}$ according to $\mathcal{P}(s_{t+1} | s_t, \bm{a_t})$ and provides a shared reward $r(s_t, \bm{a_t})$. Each agent receives an observation $z^n_t$ from $\mathcal{O}(s_t, n)$ and maintains an observation-action history $\tau^n_t$. MARL aims to learn policies $\{\pi^n\}_{n=1}^N$ that maximize expected cumulative rewards:
\begin{equation}
    J(\pi)=\mathbb{E}\left[\sum_{t=0}^{T} \gamma^{t} r(s_t, \bm{a_t})\right]
    \label{Eq: MARL objective function}
\end{equation}

This paper proposes Multi-Agent Reinforcement Learning for Single Quadruped Robot Locomotion (MASQ), which applies MARL principles to treat different parts of a single quadruped robot as independent agents, trained collaboratively using shared rewards. Specifically, this paper uses MAPPO \cite{yu2022surprising} to sovle the modeled multi-agent problem. MAPPO optimizes the following objective function in a multi-agent context:
\begin{equation}
\begin{aligned}
    L^{\text{\textit{CLIP}}}_{\text{\textit{MAPPO}}}(\theta) &= \sum_{i=1}^{n} \mathbb{E}_t \left[ \min \left( r_t^i(\theta_i) \hat{A}_t, \right. \right. \\
    & \left. \left. \text{clip}(r_t^i(\theta_i), 1 - \epsilon, 1 + \epsilon) \hat{A}_t \right) \right]
\end{aligned}
    \label{Eq: MAPPO loss}
\end{equation}
where 
\begin{equation}
    r_t^i(\theta_i) = \frac{\pi_{\theta_i}(a_t^i | o_t^i)}{\pi_{\theta_{i,\text{old}}}(a_t^i | o_t^i)}
    \label{Eq: MAPPO ratio}
\end{equation}
where $i$ denote the $i$-th agent in MARL. Each agent updates its policy parameters as follows:
\begin{equation}
    \theta_i \leftarrow \theta_i + \alpha \nabla_{\theta_i} L^{\textit{CLIP}}_i(\theta_i)
    \label{Eq: MAPPO policy parameters update}
\end{equation}

This paper uses centralized training with decentralized execution (CTDE)\cite{10.5555/3305381.3305500} to handle multi-agent learning challenges. This approach maintains stability in a constantly changing environment. In implementations, separate neural networks learn a policy ($\pi_\theta$) and a value function ($V_\psi(s)$) similar to the single-agent version. The value function helps reduce training variability and can take in additional global information. This approach leads to better final performance, faster learning speed, and improved robustness in deployment.

\subsection{MASQ Settings}
\label{Subsection: MASQ Settings}

\textbf{State Space and Observation:} The shared actor network takes a concatenated observation from four agents, each with 35 dimensions, including motor positions \( q_t \in \mathbb{R}^3 \), motor speeds \( \dot{q_t} \in \mathbb{R}^3 \), previous actions \( a_{t-1} \in \mathbb{R}^3 \) and \( a_{t} \in \mathbb{R}^3 \), a gait sequencing director \( d_t \in \mathbb{R}^1 \), projected gravity \( g_t \in \mathbb{R}^3 \), command values \( v_t^{cmd} \in \mathbb{R}^{15} \), body speeds \( v_b \in \mathbb{R}^3 \), and a one-hot encoding \( e_t \in \mathbb{R}^1 \). The total input for the actor network is \( o_t^{\text{actor}} \in \mathbb{R}^{140} \) (35x4). The architecture of the actor network consists of a normalization layer, followed by an MLP layer, a GRU layer, and another normalization layer, with the final output being the joint angle commands for each leg. Additionally, a bias is applied to the output for better precision in control. On the other hand, the critic network uses global observations and has a 73-dimensional input, including motor positions \( q_t \in \mathbb{R}^{12} \), motor speeds \( \dot{q_t} \in \mathbb{R}^{12} \), previous actions \( a_{t-1} \in \mathbb{R}^{12} \) and \( a_{t} \in \mathbb{R}^{12} \), gait directors \( d_t \in \mathbb{R}^4 \), projected gravity \( g_t \in \mathbb{R}^3 \), command values \( v_t^{cmd} \in \mathbb{R}^{15} \), and body speeds \( v_b \in \mathbb{R}^3 \). The output of the critic network consists of continuous \( V \) values \( V_t \in \mathbb{R}^4 \), which are used to calculate the advantage function.

\textbf{Action Space:} The output of the actor-network consists of continuous actions \( a_t \in \mathbb{R}^{12} \), and the system then uses these to calculate the torques for the 12 motors. Details can be found in Section \ref{Subsection: Sim-to-real}.

\textbf{Reward Function:} The reward functions in Table \ref{tab: rewards} are designed to optimize the robot's performance by encouraging desired behaviors and penalizing undesired ones. Key rewards include: \textit{tracking linear velocity}, which uses an exponential decay function to minimize velocity error; \textit{linear velocity Z} and \textit{angular velocity XY}, both penalizing unwanted motions to ensure stability; \textit{torques}, \textit{DOF velocity}, and \textit{DOF acceleration}, which promote energy efficiency and smoother movements; and \textit{collision}, which penalizes excessive contact forces. Additional rewards focus on action smoothness, accurate jumping, minimizing foot slip and impact velocities, and enhancing locomotion stability using Raibert's heuristic\cite{4307016} and foot velocity tracking.

\begin{table}[htbp]
    \centering
    \caption{Reward function}
    \label{tab: rewards}
    \setlength{\tabcolsep}{4pt} 
    \renewcommand{\arraystretch}{1.2} 
    \resizebox{0.5\textwidth}{!}{ 
    \begin{tabular}{l|l|l}
        \hline
        \multicolumn{3}{c}{\uppercase{\textbf{Reward Settings, Corresponding Equations, and Their Scales}}} \\
        \hline
        \textbf{Reward Term} & \textbf{Equation} & \textbf{Scale} \\
        \hline
        Tracking Linear Velocity & $\exp \left( -\frac{\|\mathbf{v}_{\text{cmd}} - \mathbf{v}_{b}\|^2}{\sigma_{t}} \right)$ & 1.0 \\
        Tracking Angular Velocity & $\exp \left( -\frac{\left(\mathbf{\omega}_{\text{cmd}, z} - \mathbf{\omega}_{b, z}\right)^2}{\sigma_{\text{yaw}}} \right)$ & 0.5 \\
        Linear Velocity Z & $\|\mathbf{v}_{b, z}\|^2$ & $-2 \times 10^{-2}$\\
        Angular Velocity XY & $\sum_i \|\mathbf{\omega}_{b}\|^2$ & $-1 \times 10^{-3}$ \\
        Angular Velocity Torques & $\sum_i \|\tau_i\|^2$ & $-1 \times 10^{-5}$ \\
        DOF Velocity & $\sum_i \|\mathbf{v}_{d, i}\|^2$ & $-1 \times 10^{-4}$ \\
        DOF Acceleration & $\sum_i \left( \frac{\mathbf{v}_{d, i, t} - \mathbf{v}_{d, i, t-1}}{\Delta t} \right)^2$ & $-2.5 \times 10^{-7}$ \\
        Collision & $\sum \left(1.0 \cdot (\|\mathbf{f}_{c}\| > 0.1)\right)$ & -5.0 \\
        Action Rate & $\sum \|\mathbf{a}_t - \mathbf{a}_{t-1}\|^2$ & $-1 \times 10^{-2}$ \\
        Jump & $-(h_b - h_j)^2$ & 10.0 \\
        Feet Slip & $\sum (c_f \cdot \|\mathbf{v}_{f}\|^2)$ & $-4 \times 10^{-2}$ \\
        Action Smoothness 1 & $\sum \left( \mathbf{a}_t - \mathbf{a}_{t-1} \right)^2$ & -0.1 \\
        Action Smoothness 2 & $\sum \left( \mathbf{a}_t - 2\mathbf{a}_{t-1} + \mathbf{a}_{t-2} \right)^2$ & -0.1 \\
        Feet Impact Velocity & $\sum (c_s \cdot \|\mathbf{v}_{f, p}\|^2)$ & -0.0 \\
        Raibert Heuristic & $\sum \| e_r \|^2$ & -10.0 \\
        Tracking Contacts Shaped Velocity & $\sum \left( c_d \cdot \left(1 - \exp \left(-\frac{\|\mathbf{v}_{f}\|^2}{\sigma_{gv}}\right) \right)\right)$ & 4.0 \\
        \hline
    \end{tabular}
    }
\end{table}


\subsection{Multi-agent Actor and Global Critic Networks}
\label{Subsection: Multi-agent actor networks}

In the simulation environment of Isaac Gym \cite{makoviychuk2021}, the robot receives observations and rewards to facilitate its learning. The learning process involves dividing the observations into shared and private features, and both the actor and critic networks use the rewards to train the policy.

The actor-network consists of a multi-layer perception (MLP) base and an activation layer that produces actions and their associated log probabilities. Similarly, the critic network uses an MLP base, ending in an output layer that predicts value functions. The actor-network outputs actions based on observations, while the critic network assesses the value of these actions to guide the learning process. The actions produced for the agents are processed by an actuator network\cite{doi:10.1126/scirobotics.aau5872} to simulate real-world conditions, enhancing deployment effectiveness in natural environments. After training, the trained actor-network is deployed onto the robotic dog to perform actions directly. Fig. \ref{Figure: learning process} illustrates the entire learning process.

In the context of a quadruped robot, we consider each leg as an individual agent. All four agents share a standard actor-network. Using a shared-parameter network instead of four separate actor networks helps reduce computational load and better fits the nature of a quadruped robot. Unlike typical multi-agent environments, such as StarCraft \cite{rashid2020monotonic}, where each soldier is an independent agent, the quadruped robot is a single entity with four legs symmetrically positioned around the body's center. Therefore, a shared-parameter actor-network is more suitable for this scenario.

We express the policy for each leg as follows:
\begin{equation}
    \pi_{\theta_i}(a_{i,t} \mid s_{i,t}) = \pi_{\theta}(a_{i,t} \mid s_{i,t})
    \label{Eq: Shared Parameter Network}
\end{equation}
where \text{$i$ = 1, 2, 3, 4 corresponds to the four legs.}

The quadruped robot has four legs, each with three motors: hip, thigh, and calf joint motors. Each motor's position, velocity, and torque are observable, so we use these details as the independent observations for each agent. Additionally, to enhance the coordination among the agents, we augment each independent observation with shared observations, including speed and gravity projections calculated from inertial measurement unit (IMU) data, temporal director, and agent identifier (ID). The temporal director $T_i(t)$ guides the gait sequence of each leg under different movement postures, while the agent ID is necessary for the shared-parameter network. This setup ensures the independence of each agent while improving their cooperative capabilities.\label{observation}
The temporal director helps to synchronize the movements of different legs, ensuring smooth gait patterns. It can be defined as:
\begin{equation}
    T_i(t) = \sin\left( 2\pi(k t + \Delta_i) \right)
    \label{Eq: temporal director}
\end{equation}
where
\begin{itemize}
    \item \( k \) is the scaling factor of scaling factor of gait cycle.
    \item \(\Delta_i\) is the phase offset for the \(i\)-th leg, which determines its relative timing within the gait cycle to ensure coordinated movement.
\end{itemize}

\begin{figure*}[htbp]
    \centering
    \subfloat[Robustness in outside disturbing]{  
    \begin{minipage}[b]{0.9\textwidth} 
        \centering
        \includegraphics[width=15cm,height=1.8cm]{robustness_list_compressed.png} 
    \end{minipage}}

    \subfloat[Contact force in different legs while encountering force]{ 
    \begin{minipage}[b]{0.9\textwidth} 
        \centering
        \includegraphics[width=16cm,height=2cm]{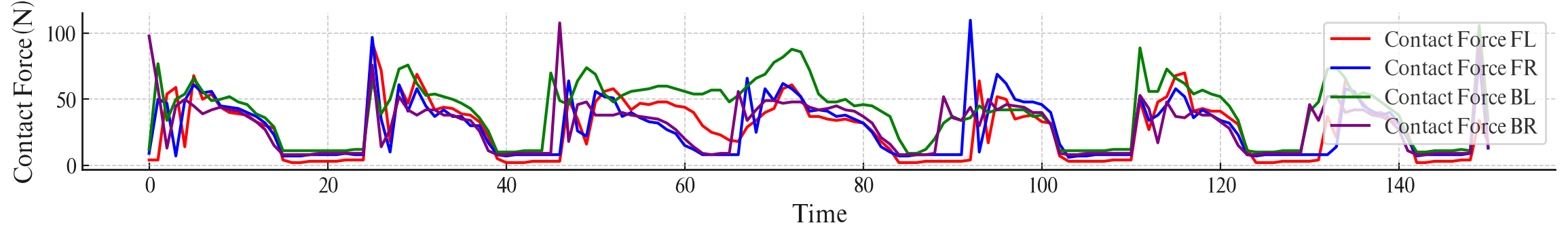} 
    \end{minipage}}
    \caption{Robustness test}
    \label{fig: robustness2} 
\end{figure*}

The Global Critic uses a centralized value function approach to consider global information, which fits into the CTDE. It relies on a global value function to coordinate individual agents, such as single-agent algorithms like PPO. The Critic network takes in global observations, which consist of the separate observations of the four agents and shared global information.

During the training in the simulation environment, we used a multi-gait curriculum learning \cite{10.1145/1553374.1553380}. This curriculum comprises four gaits: pace, trot, bound, and pronk. In the simulation, the quadruped robot learns these four gaits simultaneously, and the progress is updated based on evaluating whether the gaits' rewards meet specific thresholds. This method allows the robot to learn different gaits effectively and is also helpful in testing the generalization capabilities of our proposed approach across various tasks in experimental settings.

\subsection{Sim-to-real}
\label{Subsection: Sim-to-real}

To bridge the gap between simulation and real-world performance, we used domain randomization and an Actuator Network\cite{doi:10.1126/scirobotics.aau5872}. Domain randomization involves randomizing various parameters to train a robust policy under different conditions\cite{Peng_2018,8954361,8202133}. These parameters include robot body mass, motor strength, joint position calibration, ground friction, restitution, orientation, and magnitude of gravity. We also independently randomize friction coefficients like ground friction. Gravity is randomized every 8 seconds with a gravity impulse duration of 0.99 seconds. Time steps are randomized every 6 seconds, with the overall randomization occurring every 4 seconds. These measures enhance the model's robustness and adaptability. The training process for the Actuator Network captures the non-ideal relationship between PD error and the torque realized \cite{doi:10.1126/scirobotics.aau5872}, thereby improving the model's performance in real-world applications.

\section{Experiments}
\label{Section: Experiments}
We conducted experiments using the Unitree Go2 quadrupled robot with various experiments. Section \ref{Subsection: Experiments Setup} proposes the experimental setup. Section \ref{Subsection: Experiments in Simulation} shows the experimental results of the simulation. Section \ref{Subsection: Real-word Experiments} shows the real-world experiments and comparisons.

\begin{figure}[htbp]
    \centering
    \subfloat[Flat terrain]{  
    \begin{minipage}[b]{0.25\textwidth} 
        \centering
        \includegraphics[width=3.8cm, height=2cm]{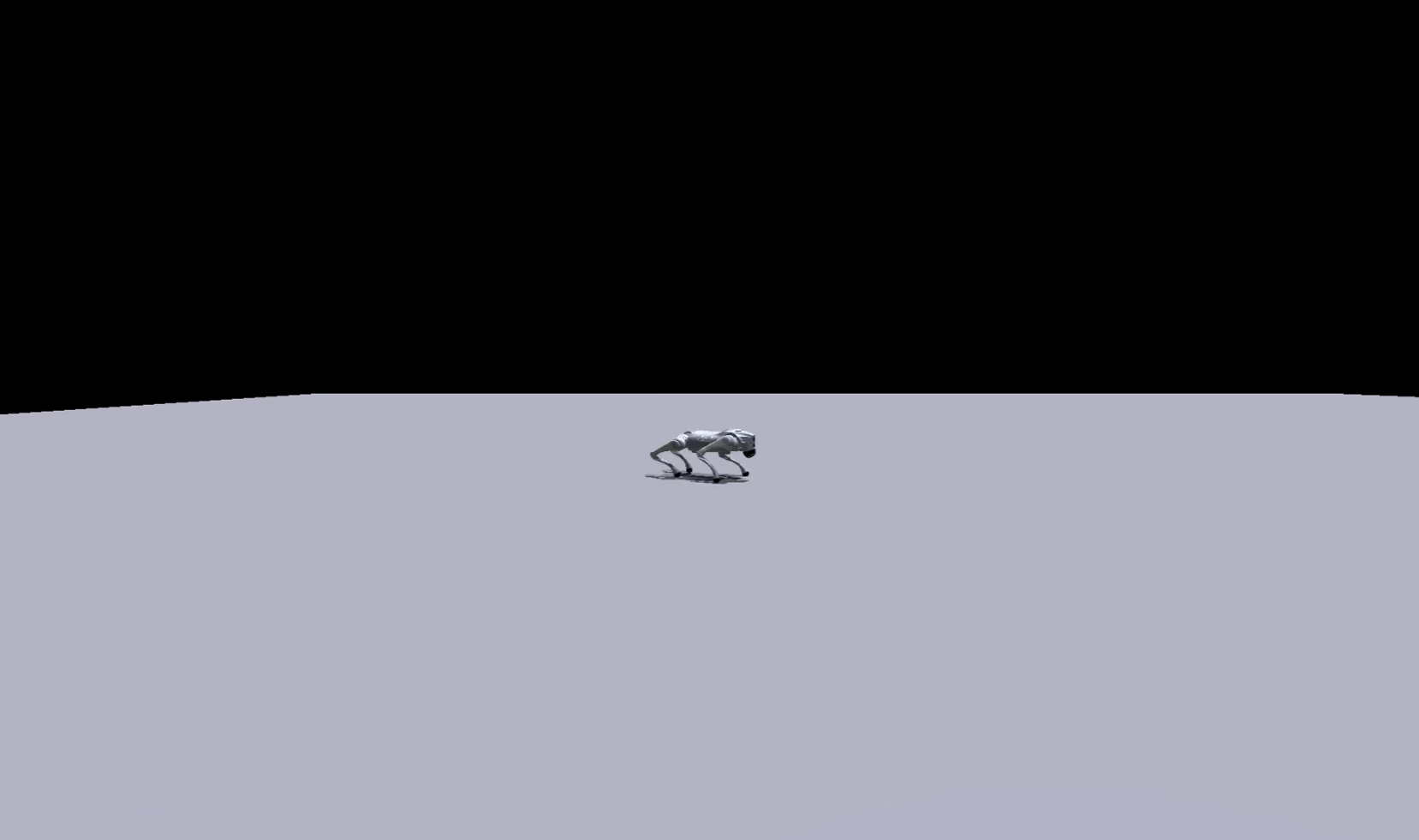} 
    \end{minipage}}
    \subfloat[Uneven terrain]{ 
    \begin{minipage}[b]{0.25\textwidth} 
        \centering
        \includegraphics[width=3.8cm, height=2cm]{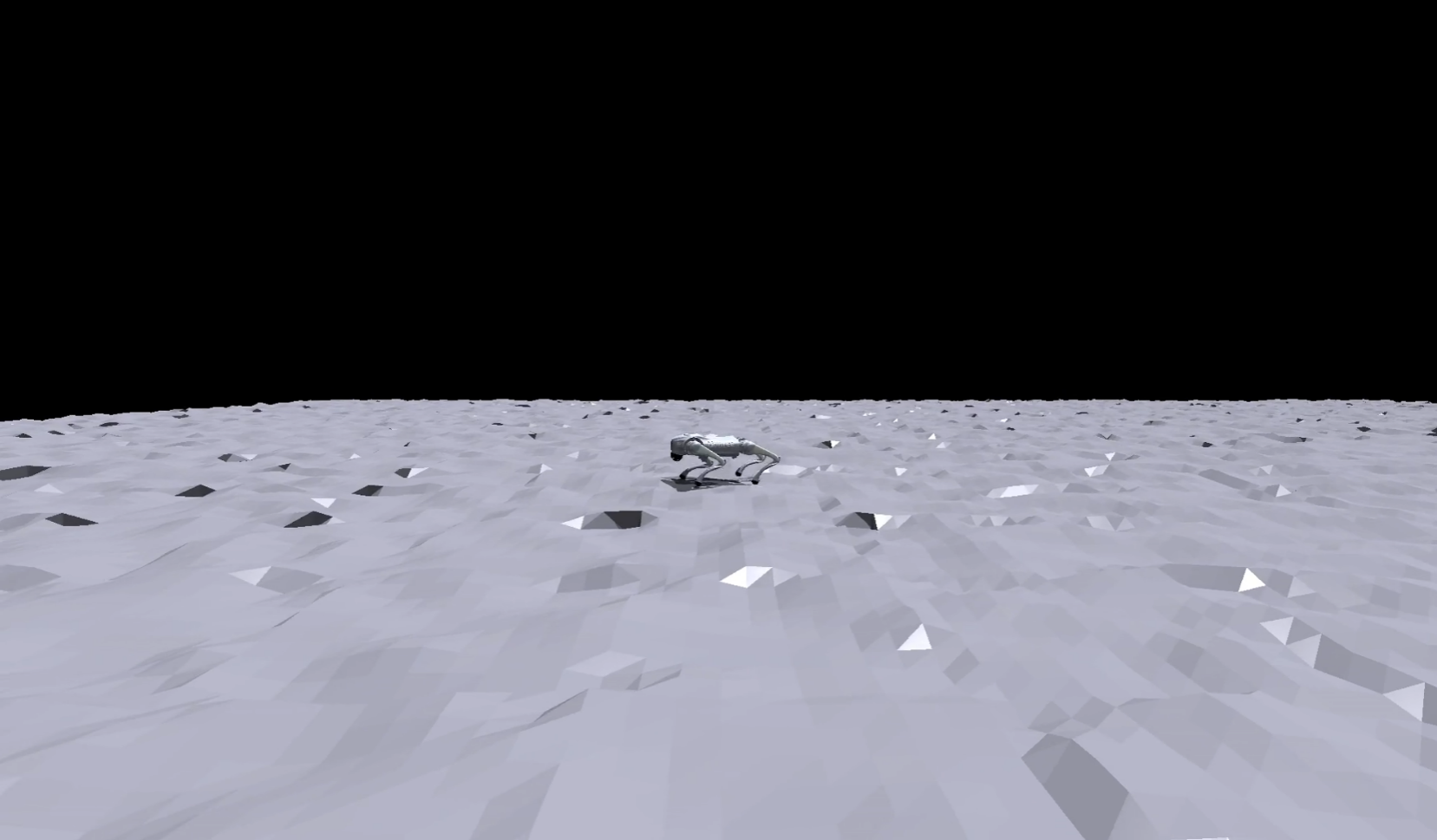} 
    \end{minipage}}
    \caption{Various terrains}
    \label{fig: Various terrains in simulation} 
\end{figure}

\subsection{Experiments Setup}
\label{Subsection: Experiments Setup}
In the simulation environment, we designed two types of terrain: flat and uneven, shown in Fig. \ref{fig: Various terrains in simulation}. Uneven terrain is a composite of pyramid-sloping terrain and random uniform terrain. Fifteen commands, such as linear velocity, angular velocity and height of the base, sampled within a specified range, guide the gait-learning process. The robot learns to adapt and develop various skills by following these commands.

\subsection{Simulation Experiments}
\label{Subsection: Experiments in Simulation}



\begin{figure}[htbp]
    \centering
    \subfloat[Reward over time (flat terrain)]{  
    \begin{minipage}[b]{0.5\textwidth} 
        \centering
        \includegraphics[width=8cm, height=4.53cm]{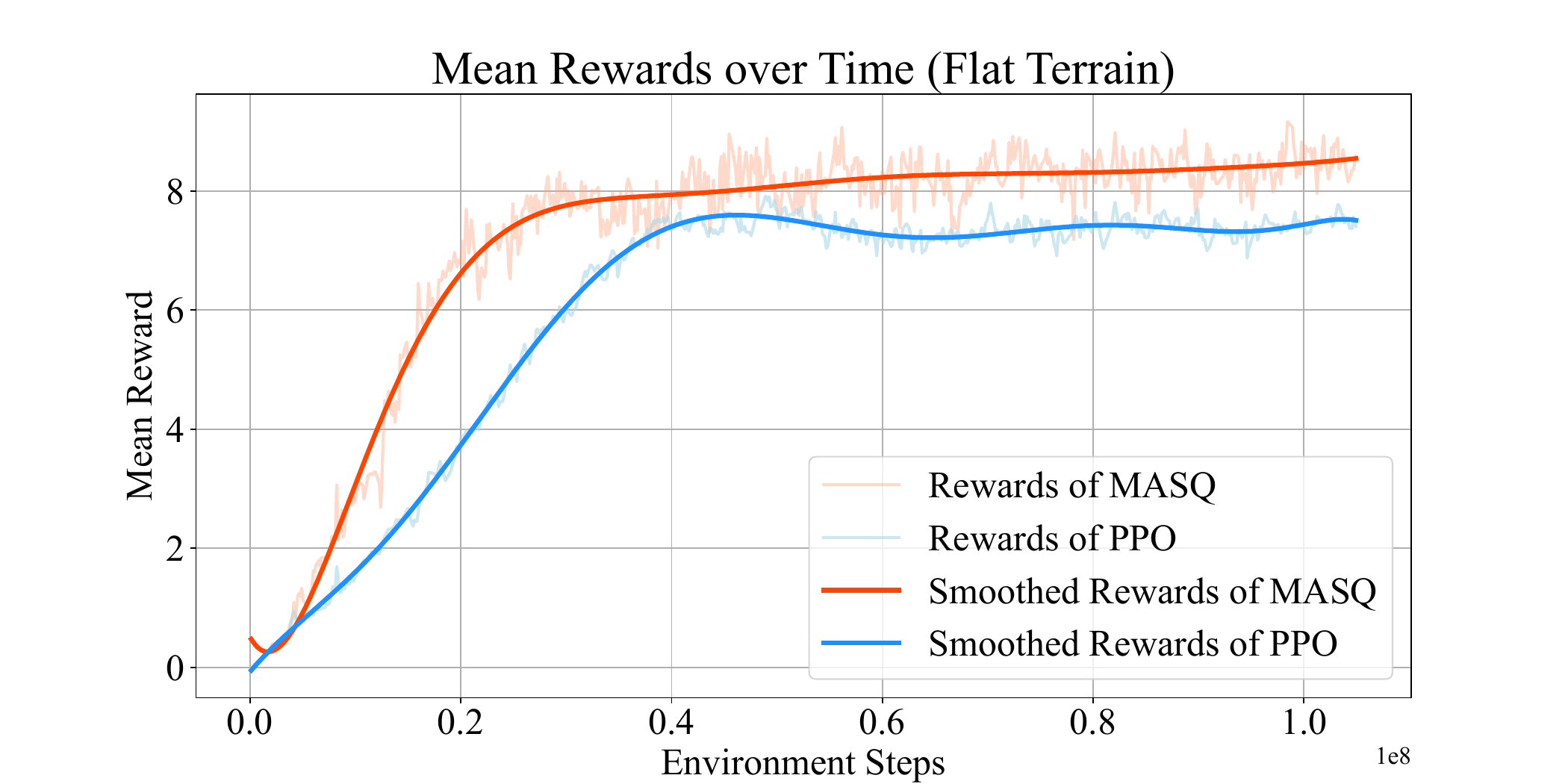} 
    \end{minipage}}

    \subfloat[Reward over time (uneven terrain)]{ 
    \begin{minipage}[b]{0.5\textwidth} 
        \centering
        \includegraphics[width=8cm, height=4.53cm]{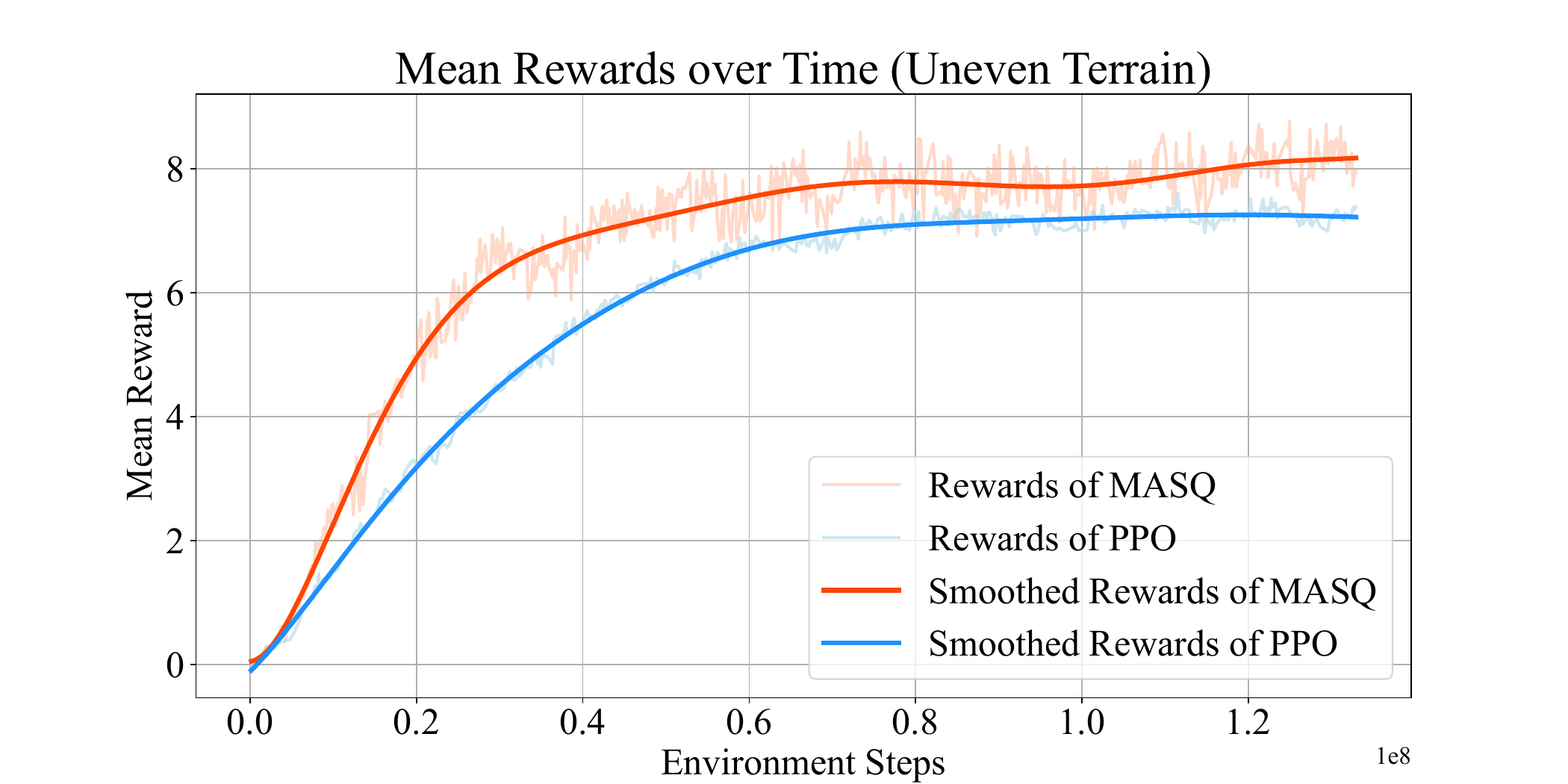} 
    \end{minipage}}
    \caption{Return on various terrains}
    \label{fig: Return} 
\end{figure}

\begin{figure}[htbp]
    \centering
    \includegraphics[width=0.75\linewidth]{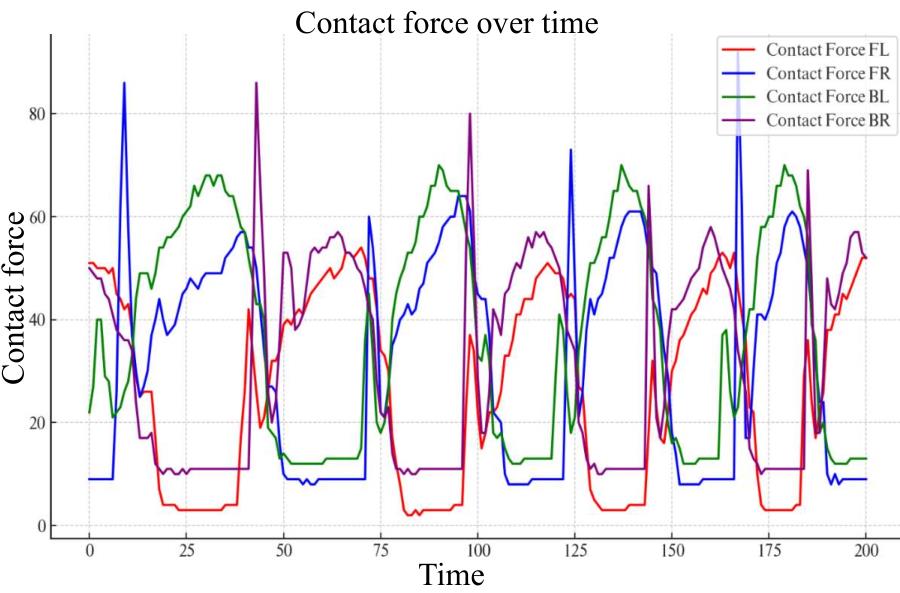}\\[1ex]
    \includegraphics[width=0.75\linewidth]{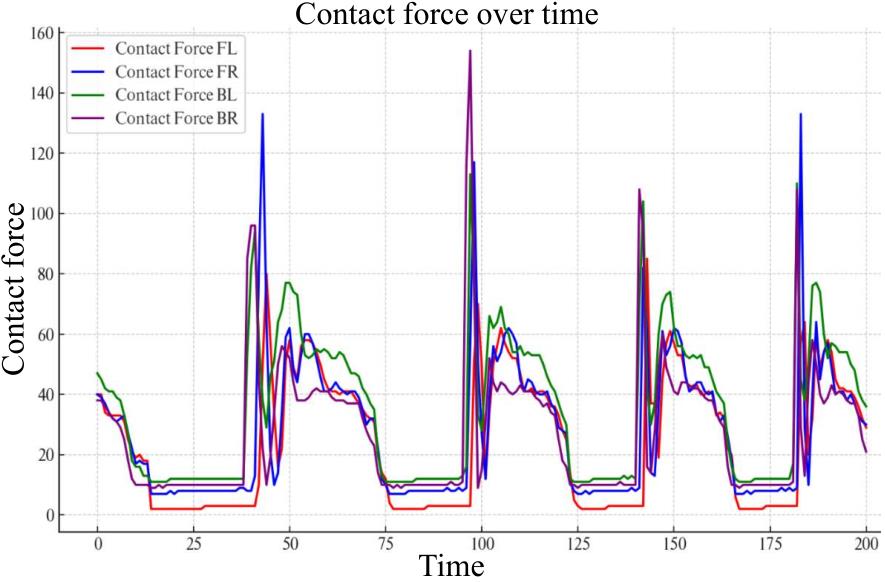}\\[1ex]
    \includegraphics[width=0.75\linewidth]{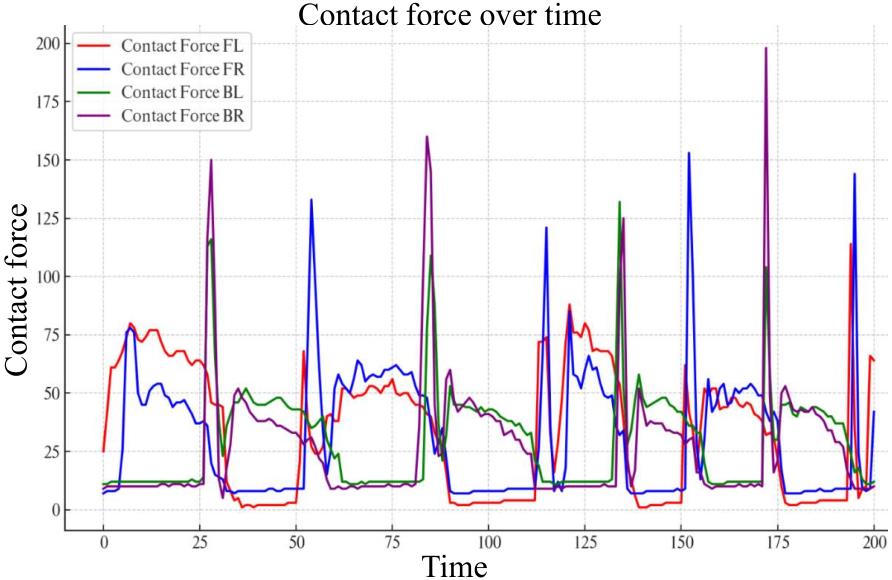}\\[1ex]
    \includegraphics[width=0.75\linewidth]{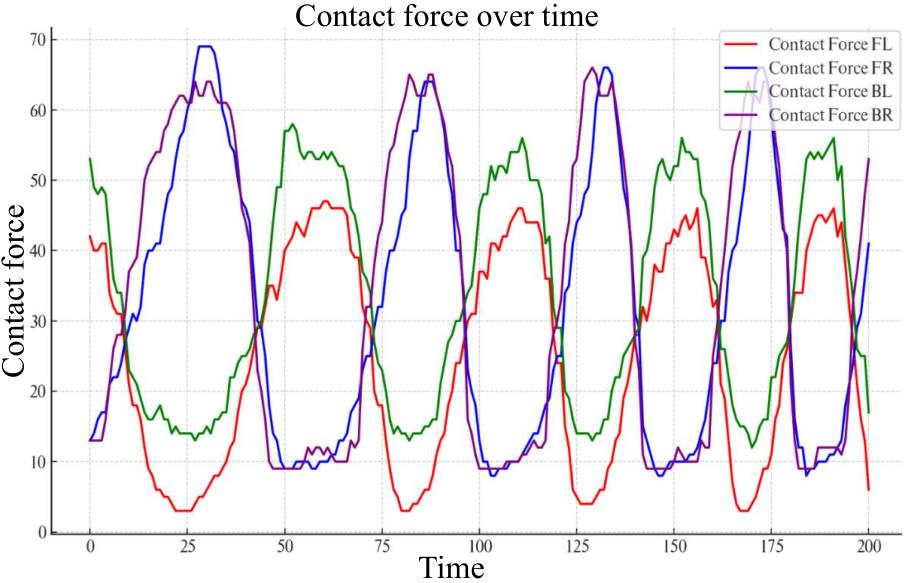}
    \caption{Contact force of four legs in different gaits}
    \label{fig: dog2}
\end{figure}

The MASQ was compared with the PPO student-teacher baseline in simulation environments, including two types: flat and uneven terrain. The quadruped robot was trained using the curriculum learning method, with hyperparameters such as the learning rate tuned for optimal performance in both scenarios. Reward curves in Fig. \ref{fig: Return} show that the proposed method achieves faster convergence and better final performance compared to the PPO baseline. Experimental results are presented as the mean of rewards over five tests for a fair comparison.

\subsection{Real-word Experiments}
\label{Subsection: Real-word Experiments}
This paper deploys MASQ on a quadruped robot and tests it on various terrains, including flat ground, grass, and sand in Fig. \ref{Figure: dog_pictures}, where it performs excellently. In addition, we conducted a series of robustness tests for heavy impacts and side kicks. The robot can quickly return to its normal state after being disturbed.

\subsubsection{\textbf{Gaits test}}
\label{Subsubsection: Gaits Test}

Fig. \ref{fig: dog2} illustrates the periodic relationship of quadruped force feedback for each of the four legs under the training of four different skills. They reflect the impact of incorporating the temporal director in our observations on the learning and switching of other skills.

\subsubsection{\textbf{Robustness test}}
\label{Subsubsection: Robustness Test}

To validate the robustness of the gaits trained with our method, we conduct external disturbance tests on the robot. We test the robot in a bounding gait with a cycle period of 20ms. During robustness tests, the robot performs continuous jumps in this bounding gait. The impacts of human steps on the robot are applied to propose disturbances during its jump cycle. Under normal conditions, the robot landed simultaneously on all four legs. We monitored this process by recording the force sensor values in the robot's feet, thereby documenting the transition from a normal state through the induced disturbance and back to the normal state. As shown in Fig. \ref{fig: robustness2}, the robot returns to its normal state after experiencing a disturbance within just one gait cycle.


\section{Conclusion and Future Work}
\label{Section: Conclusion and Future Work}

This paper proposed a multi-agent-based motion control system for quadruped robots, utilizing a shared-parameter actor network and a centralized critic network within the CTDE framework. The proposed approach, implemented in Isaac Gym, demonstrated substantial improvements in training speed, robustness, and final performance, benefiting from curriculum learning and domain randomization. These advances enabled efficient limb coordination and smoother sim-to-real transitions. Experimental results confirmed the effectiveness of the method in enhancing both performance and efficiency in motion control for symmetric robots. Future work will extend the approach to other symmetric robots and explore its application in more complex dynamic environments.

\bibliographystyle{IEEEtran}
\bibliography{ref}

\end{document}